\documentclass{llncs}

\usepackage{url}
\usepackage{cite}
\usepackage{graphics}
\usepackage{graphicx} 
\usepackage{graphics}
\usepackage[tight,footnotesize]{subfigure}
\usepackage{array}
\usepackage{amsfonts}
\usepackage{bm}
\usepackage{amsmath}
\usepackage{multirow}
\usepackage{float}

\begin{document}
\title{Segmentation of Lumen and External Elastic Laminae in Intravascular Ultrasound Images using Ultrasonic Backscattering Physics Initialized Multiscale Random Walks}



\author{Debarghya China\inst{1} \and Pabitra Mitra\inst{2} \and Debdoot Sheet\inst{3}} 

\institute{Advanced Technology Development Centre \and Department of Computer Science and Engineering \and Department of Electrical Engineering \\Indian Institute of Technology Kharagpur, India}

\maketitle

\begin{abstract}

Coronary artery disease accounts for a large number of deaths across the world and clinicians generally prefer using x-ray computed tomography or magnetic resonance imaging for localizing vascular pathologies. Interventional imaging modalities like intravascular ultrasound (IVUS) are used to adjunct diagnosis of atherosclerotic plaques in vessels, and help assess morphological state of the vessel and plaque, which play a significant role for treatment planning. Since speckle intensity in IVUS images are inherently stochastic in nature and challenge clinicians with accurate visibility of the vessel wall boundaries, it requires automation. In this paper we present a method for segmenting the lumen and external elastic laminae of the artery wall in IVUS images using random walks over a multiscale pyramid of Gaussian decomposed frames. The seeds for the random walker are initialized by supervised learning of ultrasonic backscattering and attenuation statistical mechanics from labelled training samples. We have experimentally evaluated the performance using $77$ IVUS images acquired at $40$ MHz that are available in the IVUS segmentation challenge dataset\footnote{http://www.cvc.uab.es/IVUSchallenge2011/dataset.html} to obtain a Jaccard score of $0.89 \pm 0.14$ for lumen and $0.85 \pm 0.12$ for external elastic laminae segmentation over a $10$-fold cross-validation study.

\end{abstract}


\section{Introduction}
\label{sec:intro}
Coronary artery diseases cause partial or total restriction of blood supply due to formation of hard or soft plaques that lead to a condition of constrained blood circulation around the heart. The outcome of this disorder leads to myocardial cardiac infarction (heart attack) and may even lead to death. There are several imaging techniques for \emph{in vivo} estimation of atherosclerotic plaques viz. x-ray computed tomography or magnetic resonance imaging. One of the commonly used \emph{in vivo} adjunct imaging techniques is intravascular ultrasound (IVUS) which provides detailed information about the lumen wall, compositing tissues, plaque morphology and pathology imaging is performed using a catheter is inserted in the blood vessel, with the transducer usually positioned at the tip of the catheter. A signal of high frequency (20-40 MHz) ultrasonic signal is transmitted and the beam reflected as it passed through different layers. The received signal is used for image formation and the combination of constant speed catheter’s pullback and the reflection of ultrasonic waves from the material, generates sequence of images along the length of the artery. This modality also provides information about the constituent component of plaque which clinically assist to identification of likelihood lesion of a rupture. 

\section{Prior Art}
\label{sec:priorart}
Methods for automated identification and segmentation of the lumen from media adventitia and from media externa for assessing the pathology and morphology of plaques for assisting clinicians for better diagnosis and treatment planning. Manual delineation of lumen and external elastic luminae for segmenting these layers is tedious and time consuming process. Prior art includes active surface segmentation algorithm for 3D segmentation in assessment of coronary morphology~\cite{shekhar1999three,klingensmith2000evaluation}.  Deformable shape models with energy function minimization using a hopfield neural network~\cite{plissiti2004automated}, 3D IVUS segmentation model based on the fast-marching method and using Rayleigh mixture model~\cite{cardinal2006intravascular} are amongst others. A combination of implicit anisotropic contour closing (ACC) and explicit snake model for detection of media adventitia and lumen was proposed~\cite{gil2006statistical}. A shape-driven method was proposed for segmentation of arterial wall in the rectangular domain~\cite{unal2008shape}. A knowledge based system for IVUS image segmentation was introduced which minimizes inter- and intra-observer variability~\cite{bovenkamp2009user}.  There are few IVUS border detection algorithms developed based on edge tracking and gradient based techniques~\cite{herrington1992semi,sonka1995segmentation}. A new approach based on 3-D optimal graph search was developed for IVUS image segmentation~\cite{downe2008segmentation}. A fully automated segmentation method based on graph representation was introduced for delineation of luminal and external elastic lamina surface of coronary artery~\cite{sun2013graph}. A probabilistic approach for delineation of luminal border was presented based on minimization of the probabilistic cost function~\cite{mendizabal2013segmentation}. A holistic approach for media-adventitia border detection was introduced in~\cite{ciompi2012holimab}. An automated \emph{in vivo} delineation of lumen wall had done using graph theoretic random walk method~\cite{china2016automated}.

\section{Problem Statement}
\label{sec:problem}
Let us considered an IVUS frame $I$ where $i(x)$ is the intensity at location $x$. The lumen and external elastic luminae borders split the image $I$ in to three disjoint set as $I_{lumen}$, $I_{media}$ and $I_{externa}$ such that $I_{lumen} \cap I_{media} \cap I_{externa} = \emptyset $ and $I_{lumen} \cup I_{media} \cup I_{externa} = I$. Image $I$ can be represented as an equivalent graph $\mathcal{G}$ such that the nodes of the graph can be represented as $n \in\ I$ and the edges connecting nodes of graph $\mathcal{G}$ are modeled by physics of acoustic energy propagation and attenuation within highly scatteering biological tissues. The probability of the each node $n \in\ \mathcal{G}$ to belong to either of $ \lbrace I_{lumen},\, I_{media},\, I_{externa} \rbrace $ can be solved using the random walks for image segmentation approach~\cite{roy2016lumen}. The class posterior probability at a $x \in\ I$ is the probability of the corresponging node $n \in\ \mathcal{G}$. A pixel at location $x$ is labeled as $ \arg \max \lbrace p(lumen|x,I),\, p(media|x,I),\,$ $ p(externa|x,I) \rbrace $.

A set of seeds $S$ constituting some of the marked nodes of graph $\mathcal{G}$ such that $S \subseteq \lbrace (S \in I_{lumen}) \cup (S \in I_{media}) \cup (S \in I_{externa}) \rbrace$ and $(S \in I_{lumen}) \cap (S \in I_{media}) \cap (S \in I_{externa}) = \emptyset $ is define for initialization of the random walker. Ultrasonic backscattering physics based model is used to achieve the solution of the random walker. Hence, class posterior probability would be assign to the unmarked nodes $U_m = \mathcal{G}-S$ of the graph to achieve the lumen and media adventitia border such that $\mathcal{G} \subseteq \lbrace S \cup U_m \rbrace$ and $S \cap U_m = \emptyset $.
Different stages of our proposed method have been shown in Fig.1 and that are detailed in the subsequent sections.
\begin{figure}
\centering
\includegraphics[width=0.88\textwidth]{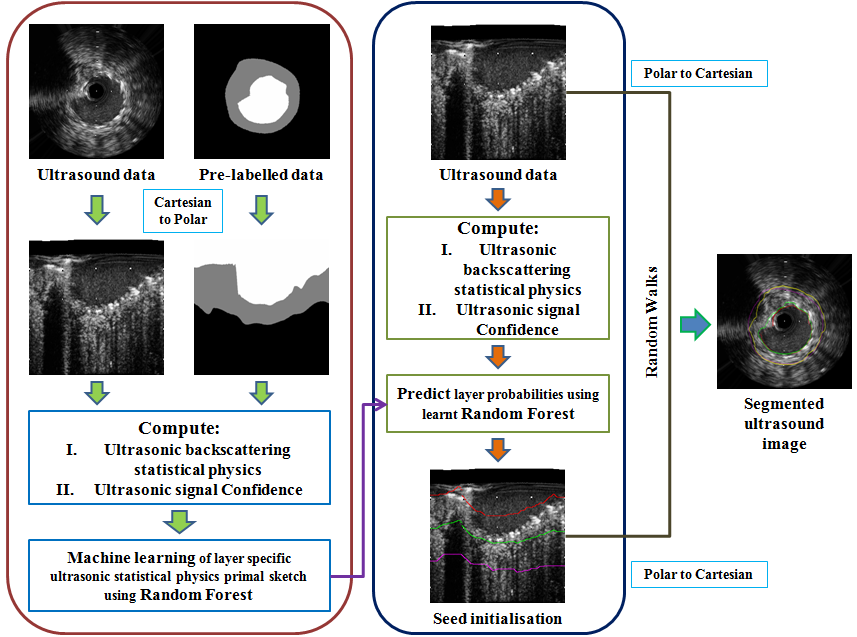}
\caption{Flow diagram of our proposed method for lumen and external elastic luminae segmentation from intravascular ultrasound images}
\end{figure}
\section{Exposition to the Solution}
\label{sec:exposition}
Ultrasonic acoustic pulse travelling through tissues are either backscattered, attenuated or absorbed. Ultrasonic images are formed using the returned echos caused by backscattering acoustic pulse. The main cause of backscattering are the scatteres present in the tissue where there nature varies with the tissue types. The statistical nature of the envelope of ultrasonic echo $(R)$ depends on the nature of the scatterers. The Scatterers' contribution is normally treated as random walk on account of their random location within the resolution limit of the range cell of the propagating ultrasonic backscattering pulse. Let $r$ be the value of the sensed signal at an instant and $R=\{r\}$ represents the set of values recorded by the transducer array, where $R$ is purely stochastic in nature. Let $y$ be a type of tissue and ${y} \in {Y}$ represents the set of tissue type. Probability of a tissue type $y$, characterized by ultrasonic echo envelop $r$, can be written in Bayesian paradigm as
\begin{equation}
p(y|r)=\frac{p(r|y)P(y)}{p(r)}
\end{equation}
where $p(r|y)$ is the conditional likelihood of received signal $r$ from a known tissue type $y$; $P(y)$ is the prior probability of tissue type $y$, and $p(r)$ is the evidence of $r$. Since there are three tissue types viz. Lumen, Media and Externa, thus ${Y} \in {\{Lumen, Media, Externa}\}$.

Let $f_1(r;\dots|y)$ be the parametric stochastic model of ultrasonic backscattering echos and $f_2(r;\dots|y)$ be the model of signal attenuation of ultrasonic propagation. Hence we can write
\begin{equation}
p(r;\dots|y)\propto \left \{ \quad  \underset{backscattering \; \: 
stats.}{
\underbrace{f_{1}(r;\phi_1|y)}} \quad , \quad \underset{signal \; \: 
attenuation}{ \underbrace{f_{2}(r;\phi_2|y)}} \quad \right \}
\end{equation}
The properties of the tissues can be denoted using statistical physics model $\Theta = \{\phi_1, \phi_2 \}$. So, each type of tissue is characterized by a unique set of statistical physics model $\{\Theta|y\} \: \forall \: {Y}$.

Accordingly $(1)$ can be re-written by taking $(2)$ into consideration
\begin{equation}
p(y|\Theta;r)=\frac{p(\Theta;r|y)P(y)}{p(\Theta;r)}
\end{equation}
where $p(y|\Theta;r)$ is the posterior probability of predicting the type of tissue $y$, and $p(\Theta;r|y)$ is the conditional likelihood of tissue specific backscattering and ultrasonic pulse propagation model as modeled in $(2)$.
In order to solve this problem we would have to $(i)$ estimate the backscattering statistical physics of ultrasonic echos, $(ii)$ estimate signal attenuation of received ultrasonic echos and $(iii)$ machine learning of the primal $(\Theta;r)$ for solution of $(3)$.

\subsection{Statistical Physics of Ultrasonic Backscattering }
\label{sec:exposition:subsec:statphysics}
The conditional distribution of the random variable ${r} \in {R}$ is Nakagami distributed~\cite{shankar2000general} such that $p(r|y)\propto\ N(r|m,\Omega)$ represented as
\begin{equation}
N(r|m,\Omega) = \frac{2m^mr^{2m-1}}{\Gamma(m)\Omega^m} exp(-\frac{m}{\Omega}r^2)U(r)
\end{equation}
where $m$ and $\Omega$ are known as the Nakagami shape and scale factors respectively. $\Gamma(.)$ is the mathematical Gamma distribution function and $U(.)$ is the unit step response. The parameters are estimated from the moments of the enveloped signal $R$ as
\begin{equation}
m = \frac{(E[R^2])^2}{E[R^2-E[R^2]]^2} \quad \textrm{and} \quad
\Omega = E[R^2]
\end{equation}
where $E(.)$ is the mathematical expectation operator. Since in a B-mode image, the image intensity $i$ is a log-compressed version of the signal $r$, the intensity ${i} \in {I}$ is accordingly Fisher-Tippett distributed~\cite{shankar2003estimation,sheet2013random} such that $p(i|y) \propto\ F(i|\sigma)$ and
\begin{equation}
F(i|\sigma) \propto\ exp([2i-ln(2\sigma^2)]-exp[2i-ln(2\sigma^2)])
\end{equation}
where $\sigma$ is the standard deviation of intensity~\cite{shankar2003estimation}.

The parameters of ${i}$ and ${\sigma}$ are estimated through a nonlinear multiscale estimation. According to our proposition, these parameter are estimated at different scales $\tau = (\tau_{trans}, \tau_{axial})$ where $\tau_{trans}$ is the number of neighboring scan lines and $\tau_{axial}$ is the number of samples along each scan line, with $(\tau_{trans}, \tau_{axial}) \in \lbrace (3,3),(3,5),(3,7),...,(3,30)\rbrace $ such that estimation holds true for the strong law of large numbers\cite{sen1994large}. Thus an ordered vector of ${(i,\sigma)} \subset \Theta$ forms part of the information required for solving $(3)$.

\subsection{Ultrasonic Signal Confidence Estimation}
\label{sec:exposition:subsec:signalconfidence}
Ultrasonic signal attenuation measured as signal confidence in $(3)$ is estimated using the method of random walks~\cite{sheet2014joint,karamalis2012confidence,karamalis2012ultrasound}. The backscattered ultrasonic echos from randomly distributed scatterers can be treated as a random walk walking from a point in space to the transducer~\cite{shankar2000general}. According to that concept, backscattered echo and ultrasonic pulse are travelling through the same path of a heterogeneous media are subjected to the same attenuation. The confidence of the ultrasonic signal has been estimated as the probability of a random walker starting at a node on the scan-line and reaching to the origin of each scan line where the virtual transducer element placed. Thus the signal confidence is represented as 
\begin{equation}
p(r;...|y) \propto\ f_2(r;\phi_2|y)
\end{equation}
where $\phi_2$ is the received ultrasonic signal confidence associated with backscattered echo $r$ by a tissue type $y$.

\subsection{Learning of Statistical Mechanics of Ultrasonic Backscattering for Initial Seed Selection}
\label{sec:random forest}
The parameters of $f_1(r;...|y)$ and $f_2(r;...|y)$ constitute subspaces of jointly model for prediction of the tissue specific posterior probability. Non-parametric machine learning framework of random forest~\cite{breiman2001random,criminisi2012decision} has been employed for this purpose.The prediction model of random forest can be represented as
\begin{equation}
p(y|\Theta;r) = H(y|\Theta;r)
\end{equation}
where $H(.)$ is the learnt random forest model. A random forest $H(y|\Theta;r)$ is formally defined as a classifier consisting of a collection of tree-structured decision maker $\{h(y|\Theta,\Phi_k),k=1,...\}$ where $\{\Phi_k\}$ are indepedent identically distributed random vectors which represent sample features and each tree $h(y|\Theta,\Phi_k)$ casts a unit vote for the most popular class $y$ at input $\Theta$~\cite{breiman2001random}. At the time of learning of the model, each tree is a binary tree and trained on the independent random vector $\Phi_k$. There are $nTrees \in\ \mathbb{N}$ number of trees constituting the forest model. We employ $nPercentToSample \in\ \mathbb{R}_+$ random samplingwith replacement from complete observation space for generating $\Phi_k$. Depending on the response of the $weakLearner$ each of the nodes split to two children. If the number of observation at the node is less than $minLeaf \in\ \mathbb{N}$ or if all the observations at the node belong to a single class then the splitting test on a node stop. During prediction, the vote casted by the forest is the class specific mean response of each of the trees such that $p(y|\Theta;i) = E[h(...,\Phi_k)]$. Initial segmentation is done using this random forest model to obtain seeds for each tissue type $Y \in \{lumen, media, externa\}$. This initial labels are considered as the seed points of those individual labels. Final label of segmentation has been done with graph theoretic random walk method using the initial label seed points.

\subsection{Random Walks for Lumen and External Elastic Luminae Segmentation}
\label{sec:random walks}
The graph $G$ is represented as a combinatorial Laplacian matrix $L$ for achieving an analytically convergent solution~\cite{grady2006random,roy2016lumen}.
\begin{equation}
L_{pq} = \left\{\begin{matrix}
d_p & \textrm{ if } p=q\\ 
 -w_{pq} & \quad \quad \textrm{ if } v_p \textrm{ and } v_q \textrm{ are adjacent nodes}\\
 0 & \textrm{ otherwise}
\end{matrix}\right.
\end{equation}
where $L_{pq}$ is indexed by vertices $v_p$ and $v_q$. The set of vertices or nodes $V$ can be divided into two, $V_M$ consisting of marked seeded nodes and $V_U$ consisting of unmarked or unseeded nodes such that $V_M \cup V_U = V$ and $V_M \cap V_U = \emptyset$.Thus the Laplacian matrix can be decomposed as 
\begin{equation}
L = \begin{bmatrix}
    L_M & B \\
    B^T & L_U
   \end{bmatrix}
\end{equation}
where $L_M$ and $L_U$ are Laplacian submatrices corresponding to $V_M$ and $V_U$, respectively. We denote the probability of a random walker starting at a node $v_q$ to reach a seeded point belonging to tissue type $\omega \in \{lumen, media, externa\}$ as $x_q^w$ s.t. $\sum_{\omega} x_q^w = 1$. Further, to achieve a solution, the set of labels defined for all the seeds in $ V_M \in S$ is specified using a function
\begin{equation}
Q(v_q) = \omega \quad\ and \quad \forall v_q \in V_M
\end{equation}
where $\omega \in \mathbb{Z}, 0<\omega<3\ $ s.t.$\ \omega = 1$ is the set of label corresponding to $I_{lumen}$, $\omega = 2$ is the set of labels corresponding to $I_{media}$, and $\omega = 3$ is the set of labels corresponding to $I_{externa}$. This helps us in defining $M \in S$ is a 1-D vector of $|V_M|\times1$ elements corresponding to each label at a node $v_q \in V_M$ constituted as 
\begin{equation}
m_q^\omega = \left\{\begin{matrix}
1 & \textrm{ if } Q\left(v_q\right)=\omega\\ 
 0 & \textrm{ if } Q\left(v_q\right)\neq \omega
\end{matrix}\right.
\end{equation}
Therefore, for label $\omega$, the solution can be obtained by solving
\begin{equation}
L_U x_q^\omega = -B^T m_q^\omega
\end{equation}
\begin{equation}
L_U X = -B^T M
\end{equation}
where solving for $\omega = 1$ yields $X = \lbrace x_q \forall q|v_q \in V \rbrace$ as the set of solution probabilities of a random walker originating at a node $q \in G$ and reaching the lumen and is associated and solved accordingly
\begin{equation}
p(lumen|x,I) = x_q^\omega \forall \lbrace q \in G \Leftrightarrow x \in I \rbrace, \omega = 1
\end{equation}
\begin{equation}
p(media|x,I) = x_q^\omega \forall \lbrace q \in G \Leftrightarrow x \in I \rbrace, \omega = 2
\end{equation}
\begin{equation}
p(externa|x,I) = x_q^\omega \forall \lbrace q \in G \Leftrightarrow x \in I \rbrace, \omega = 3
\end{equation}
\section{Experiments, Results and Discussions}
\label{sec:expt}
The data used in this experiment is acquired from the Lumen + External Elastic Laminae (Vessel Inner and Outer Wall) Border Detection in IVUS Challenge dataset~\footnote{http://www.cvc.uab.es/IVUSchallenge2011/dataset.html}. There are two dataset (dataset \textit{\textbf{A}} and dataset \textit{\textbf{B}}) where we only took dataset \textit{\textbf{A}} for this experiment. The data set A is composed of $77$ groups of five consecutive frames, obtained from a $40$ MHz IVUS scanner, acquired from different patients. Manually labeled data was also provided with this set of data and a MATLAB script for evaluating the results in a unified way. At the time of random forest learning, the $D$ dimentional ordered vector $\Theta$ consisting of multiscale estimated Fisher-Tippett parameter and ultrasonic signal confidence is computed and represented as $ \lbrace (\Theta;r)\forall r \in \mathcal{G} \rbrace $. In this experiments we have tissue specific labels $y \in Y$ and $Y = \lbrace lumen,media,externa \rbrace $ corresponding to the ultrasound echo measurements at a grid points $r \in \mathcal{G}$. We train a random forest model $H(y|\Theta;r)$ using the ordered vector in training case $ \lbrace (\Theta;r)\forall r \in \mathcal{G} \rbrace $. The random forest parameter is used in this work are $nTrees$ is $50$ where $treeDepth$ is $\infty$, $minLeaf$ is $50$ and $splitObj$ is Gini Diversity Index (GDI) maximization. This trained model is finally tested on selected number unknown test images(not used during training). This experiments is performed using a k-fold cross-validation technique. In this perticular experiment we have considered 10 fold cross-validation where $9$ sets of images are used for training the model and remaining one set is used for testing. The random forest is trained using $500$ randomly drawn samples for the each tissue type from an image.

After testing of the unkhown images, the classified data would taken as initial segmented layer and finally random walks algorithm has been employed for calculation of layer specific prosterior probability. Using this $10$-folded cross-validation experiments we obtain an average Jaccard coefficient value for lumen is $0.89 \pm 0.14$ and for external elastic luminae is $0.85 \pm 0.12$, the Hausdorff distances for lumen is $0.81 \pm 0.53$ and for external elastic luminae $0.95 \pm 0.48$ and the percentage of area difference value is $0.12 \pm 0.10$ for lumen and $0.10 \pm 0.09$ for external elastic luminae. The scores are competitively better from the prior art of the challenge~\cite{balocco2014standardized}. The results are approximately same as the inter- and intra-observer~\cite{balocco2014standardized}. In Table 1, the comparison is clearly drawn with the existing approaches. \textit{Appr1} shows the performance of initially segmented layer obtained from sec.4.3. Fig. 2. shows that, in the left side in Fig. 2.$(a), (c), (e)$ are shows three different sizes of lumen on the right side in Fig. 2. $(b), (d), (f)$ shows three different size of media adventitia.

\begin{table}[]
\centering
\caption{Performance evaluation metric with the dataset and comparisin with existing approaches. \textit{\textbf{P1}} and \textit{\textbf{P4}} has no experimental results with dataset \textit{\textbf{A}} in the challenge}
\scriptsize
\label{my-label}
\begin{tabular}{|c|c|c|c|c|c|c|}
\hline
\multirow{2}{*}{\textbf{Methods}} & \multicolumn{2}{c|}{\textbf{JCC}}                 & \multicolumn{2}{c|}{\textbf{HD}}                  & \multicolumn{2}{c|}{\textbf{PAD}}                 \\ \cline{2-7} 
                                  & \textit{\textbf{Lumen}} & \textit{\textbf{Media}} & \textit{\textbf{Lumen}} & \textit{\textbf{Media}} & \textit{\textbf{Lumen}} & \textit{\textbf{Media}} \\ \hline
\textit{\textbf{P2}}~\cite{balocco2014standardized}                       & { $0.75\pm 0.11$ }   & -          & { $1.78\pm 1.13$ } & -          & { $0.19\pm 0.12$ }  & -          \\ \hline
\textit{\textbf{P3}}~\cite{balocco2014standardized}                       & { $0.85\pm 0.12$ }  & { $0.86\pm 0.11$ } & { $1.16\pm 1.12$ } & { $1.18\pm 1.02$ } & { $0.10\pm 0.12$ }  & { $0.10\pm 0.11$ } \\ \hline
\textit{\textbf{P5}}~\cite{balocco2014standardized}                       & { $0.72\pm 0.12$ }  & -          & { $1.70\pm 1.09$ } & -          & { $0.22\pm 0.14$ }  & -          \\ \hline
\textit{\textbf{P6}}~\cite{balocco2014standardized}                       & -           & { $0.76\pm 0.11$ } & -          & { $1.78\pm 0.83$ } & -           & { $0.17\pm 0.14$ } \\ \hline
\textit{\textbf{P7}}~\cite{balocco2014standardized}                       & { $0.83\pm 0.12$ }  & -          & { $1.20\pm 1.03$ } & -          & { $0.14\pm 0.17$ }  & -          \\ \hline
\textit{\textbf{P8}}~\cite{balocco2014standardized}                       & { $0.80\pm 0.14$ }  & { $0.80\pm 0.13$ } & { $1.32\pm 1.18$ } & { $1.57\pm 1.03$ } & { $0.11\pm 0.12$ }  & { $0.14\pm 0.16$ } \\ \hline
\textit{\textbf{Intra-obs}}~\cite{balocco2014standardized}                & { $0.86\pm 0.10$ }  & { $0.87\pm 0.11$ } & { $1.04\pm 0.95$ } & { $1.14\pm 1.00$ } & { $0.10\pm 0.10$ }  & { $0.11\pm 0.14$ } \\ \hline
\textit{\textbf{Inter-obs}}~\cite{balocco2014standardized}                & { $0.92\pm 0.06$ }  & { $0.91\pm 0.07$ } & { $0.67\pm 0.52$ } & { $0.85\pm 0.60$ } & { $0.05\pm 0.06$ }  & { $0.06\pm 0.07$ } \\ \hline
\textit{\textbf{Appr1}}\footnotemark        & {  $\textbf{0.72}\pm \textbf{0.18}$ } & {  $\textbf{0.69}\pm \textbf{0.13}$ } & {  $\textbf{1.35}\pm \textbf{1.14}$ } & {  $\textbf{1.49}\pm \textbf{1.21}$ } & {  $\textbf{0.26}\pm \textbf{0.19}$ } & {  $\textbf{0.29}\pm \textbf{0.17}$ } \\ \hline
\textit{\textbf{Appr2}}                    & {  $\textbf{0.89}\pm \textbf{0.14}$ } & {  $\textbf{0.85}\pm \textbf{0.12}$ } & {  $\textbf{0.81}\pm \textbf{0.53}$ } & {  $\textbf{0.95}\pm \textbf{0.48}$ } & {  $\textbf{0.12}\pm \textbf{0.10}$ } & {  $\textbf{0.10}\pm \textbf{0.09}$ } \\ \hline
\end{tabular}
\end{table}

\normalsize

\footnotetext{initially segmentation using random forest}
\begin{figure}
\centering
\includegraphics[width=1\textwidth]{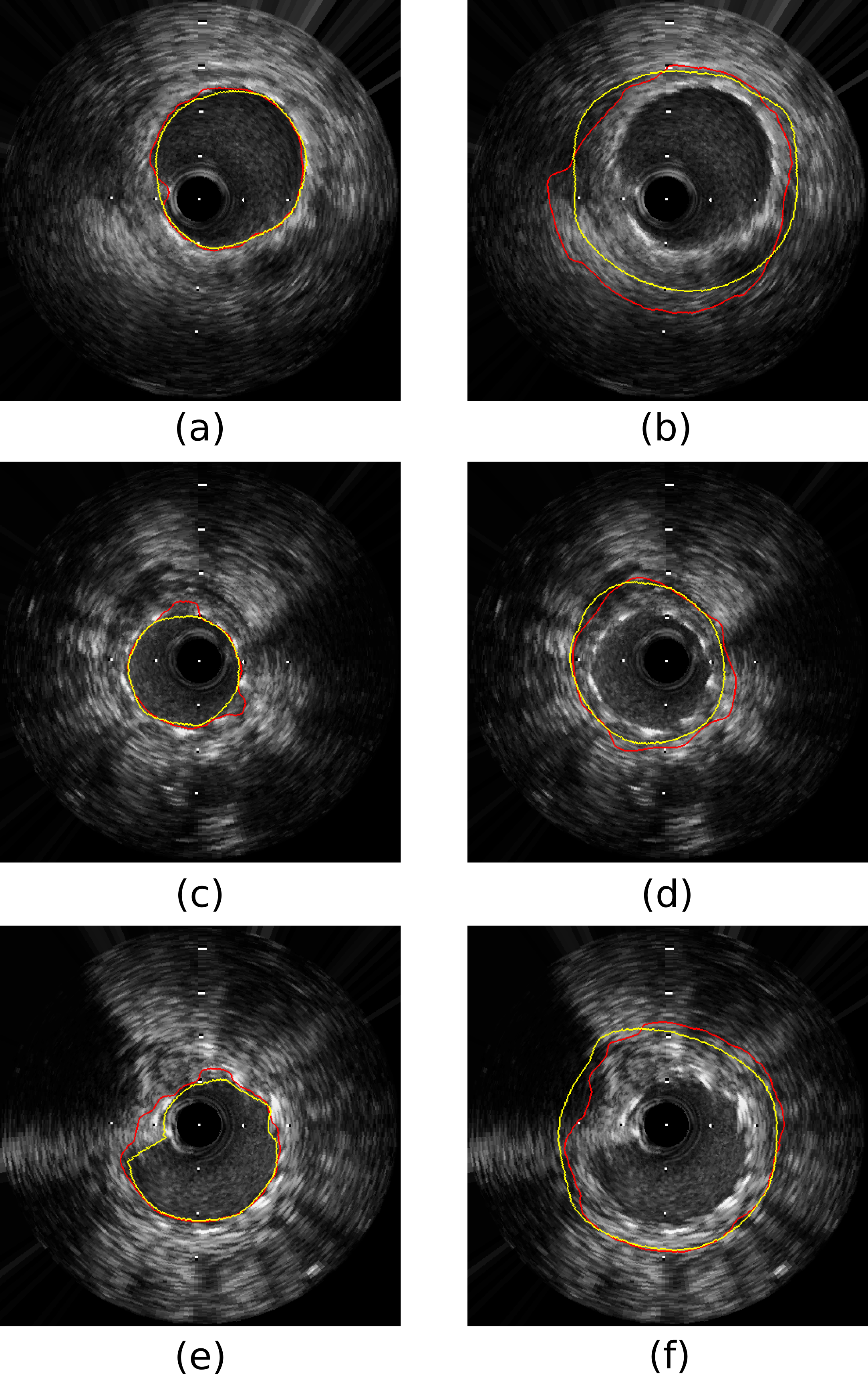}
\caption{First column is the lumen segmentation and second column is the external elastic laminae segmentation where YELLOW - ground truth data generated by the experts and RED - result of our proposed method}
\end{figure}

\section{Conclusion}
\label{sec:conc}
The approach to layer characterization or boundary detection using ultrasonic backscattered signal from the heterogenous atherosclerotic tissue is a very crucial task. In this paper, we have proposed a method for lumen and external elastic laminae segmentation in IVUS using: (i) ultrasonic backscattering physics and signal confidence estimation (ii) joint learning of these estimatesusing technique like random forests for initial layer localization and (iii) employing random walks for fine segmentation of boundaries. Our proposed algorithm is competitively more accurate and less time consuming than prior art.

\bibliographystyle{splncs03}
\bibliography{myref}

\end{document}